\documentclass[sigconf]{acmart}

\AtBeginDocument{%
  \providecommand\BibTeX{{%
    \normalfont B\kern-0.5em{\scshape i\kern-0.25em b}\kern-0.8em\TeX}}}

\copyrightyear{2024}
\acmYear{2024}
\setcopyright{acmlicensed}
\acmConference[WWW '24 Companion] {Companion Proceedings of the ACM Web Conference 2024}{May 13--17, 2024}{Singapore, Singapore.}
\acmBooktitle{Companion Proceedings of the ACM Web Conference 2024 (WWW '24 Companion), May 13--17, 2024, Singapore, Singapore}
\acmISBN{979-8-4007-0172-6/24/05}
\acmDOI{10.1145/3589335.3651572}
\usepackage{xspace}
\usepackage{multirow}
\usepackage{array}
\usepackage{xcolor}

\newcommand{\bptree}{Thought Graph\xspace}

\newcommand{\system}{Thought Graph\xspace}

\begin{document}

%%
%% The "title" command has an optional parameter,
%% allowing the author to define a "short title" to be used in page headers.
\title{\system: Generating Thought Process for Biological Reasoning}

%%
%% The "author" command and its associated commands are used to define
%% the authors and their affiliations.
%% Of note is the shared affiliation of the first two authors, and the
%% "authornote" and "authornotemark" commands
%% used to denote shared contribution to the research.

\author{Chi-Yang Hsu}
% \author{Kyle Cox}

\affiliation{%
  \institution{University of Texas at Austin}
  \city{Austin}
  \state{Texas}
  \country{USA}
}
\email{ch52669@utexas.edu}
% \email{kylecox@utexas.edu}
% \email{kylecox@utexas.edu}
% \orcid{1234-5678-9012}
\author{Kyle Cox}
\affiliation{%
  \institution{University of Texas at Austin}
  \city{Austin}
  \state{Texas}
  \country{USA}
}
\email{kylecox@utexas.edu}

\author{Jiawei Xu}
\affiliation{%
  \institution{University of Texas at Austin}
  % \streetaddress{P.O. Box 1212}
  \city{Austin}
  \state{Texas}
  \country{USA}
  % \postcode{43017-6221}
}
% \authornotemark[1]
\email{jiaweixu@utexas.edu}

\author{Zhen Tan}
\affiliation{%
  \institution{Arizona State University}
  % \streetaddress{P.O. Box 1212}
  \city{Tempe}
  \state{Arizona}
  \country{USA}
  % \postcode{43017-6221}
}
% \authornotemark[1]
\email{ztan36@asu.edu}

% \ethan{ztan36@asu.edu}

\author{Tianhua Zhai}
\affiliation{%
  \institution{University of Pennsylvania}
  % \streetaddress{P.O. Box 1212}
  \city{Philadelphia}
  \state{Pennsylvania}
  \country{USA}
  % \postcode{43017-6221}
}
% \authornotemark[1]
\email{tianhua.zhai@pennmedicine.upenn.edu}

\author{Mengzhou Hu}
\affiliation{%
  \institution{University of California San Diego}
  % \streetaddress{P.O. Box 1212}
  \city{La Jolla}
  \state{California}
  \country{USA}
  % \postcode{43017-6221}
}
\email{mhu@health.ucsd.edu}

\author{Dexter Pratt}
\affiliation{%
  \institution{University of California San Diego}
  % \streetaddress{P.O. Box 1212}
  \city{La Jolla}
  \state{California}
  \country{USA}
  % \postcode{43017-6221}
}
\email{depratt@health.ucsd.edu}

% \author{Dexter Pratt}
% \affiliation{%
%   \institution{University of California San Diego}
%   % \streetaddress{P.O. Box 1212}
%   \city{La Jolla}
%   \state{California}
%   \country{USA}
%   % \postcode{43017-6221}
% }
% \email{depratt@health.ucsd.edu}

\author{Tianlong Chen}
\affiliation{%
  \institution{The University of North Carolina at Chapel Hill}
  % \streetaddress{P.O. Box 1212}
  \city{Chapel Hill}
  \state{North Carolina}
  \country{USA}
  % \postcode{43017-6221}
}
% \authornotemark[1]
\email{tianlong@cs.unc.edu}

\author{Ziniu Hu}
\affiliation{%
  \institution{California Institute of Technology
}
  % \streetaddress{P.O. Box 1212}
  \city{Pasadena}
  \state{California}
  \country{USA}
  % \postcode{43017-6221}
}
% \authornotemark[1]
\email{acbull@caltech.edu}

\author{Ying Ding}
\affiliation{%
  \institution{University of Texas at Austin}
  % \streetaddress{P.O. Box 1212}
  \city{Austin}
  \state{Texas}
  \country{USA}
  % \postcode{43017-6221}
}
\email{ying.ding@ischool.utexas.edu}

% \ethan{}

%%
%% By default, the full list of authors will be used in the page
%% headers. Often, this list is too long, and will overlap
%% other information printed in the page headers. This command allows
%% the author to define a more concise list
%% of authors' names for this purpose.
\renewcommand{\shortauthors}{Hsu, et al.}

%%
%% The abstract is a short summary of the work to be presented in the
%% article.
\begin{abstract}
We present the \system as a novel framework to support complex reasoning and use gene set analysis as an example to uncover semantic relationships between biological processes. Our framework stands out for its ability to provide a deeper understanding of gene sets, significantly surpassing GSEA by 40.28\% and LLM baselines by 5.38\% based on cosine similarity to human annotations. Our analysis further provides insights into future directions of biological processes naming, and implications for bioinformatics and precision medicine. Here's our 
{\href{https://github.com/ethan5437/thought-graph-www/}{\textcolor{blue}{Github Code}}}.

\end{abstract}

%%
%% The code below is generated by the tool at http://dl.acm.org/ccs.cfm.
%% Please copy and paste the code instead of the example below.
%%

% \ccsdesc[500]{Semantics and Knowledge~Semantic webs in bioinformatics}
% \ccsdesc[100]{Do Not Use This Code~Generate the Correct Terms for Your Paper}

%%
%% Keywords. The author(s) should pick words that accurately describe
%% the work being presented. Separate the keywords with commas.
% \keywords{natural language processing, biological process,}

%% A "teaser" image appears between the author and affiliation
%% information and the body of the document, and typically spans the
%% page.

% \begin{teaserfigure}
%   \includegraphics[width=\textwidth]{sampleteaser}
%   \caption{Seattle Mariners at Spring Training, 2010.}
%   \Description{Enjoying the baseball game from the third-base
%   seats. Ichiro Suzuki preparing to bat.}
%   \label{fig:teaser}
% \end{teaserfigure}

% \received{20 February 2007}
% \received[revised]{12 March 2009}
% \received[accepted]{5 June 2009}

%%
%% This command processes the author and affiliation and title
%% information and builds the first part of the formatted document.
\maketitle
\begin{figure*}[t]
    \centering
    \includegraphics[width=\textwidth]{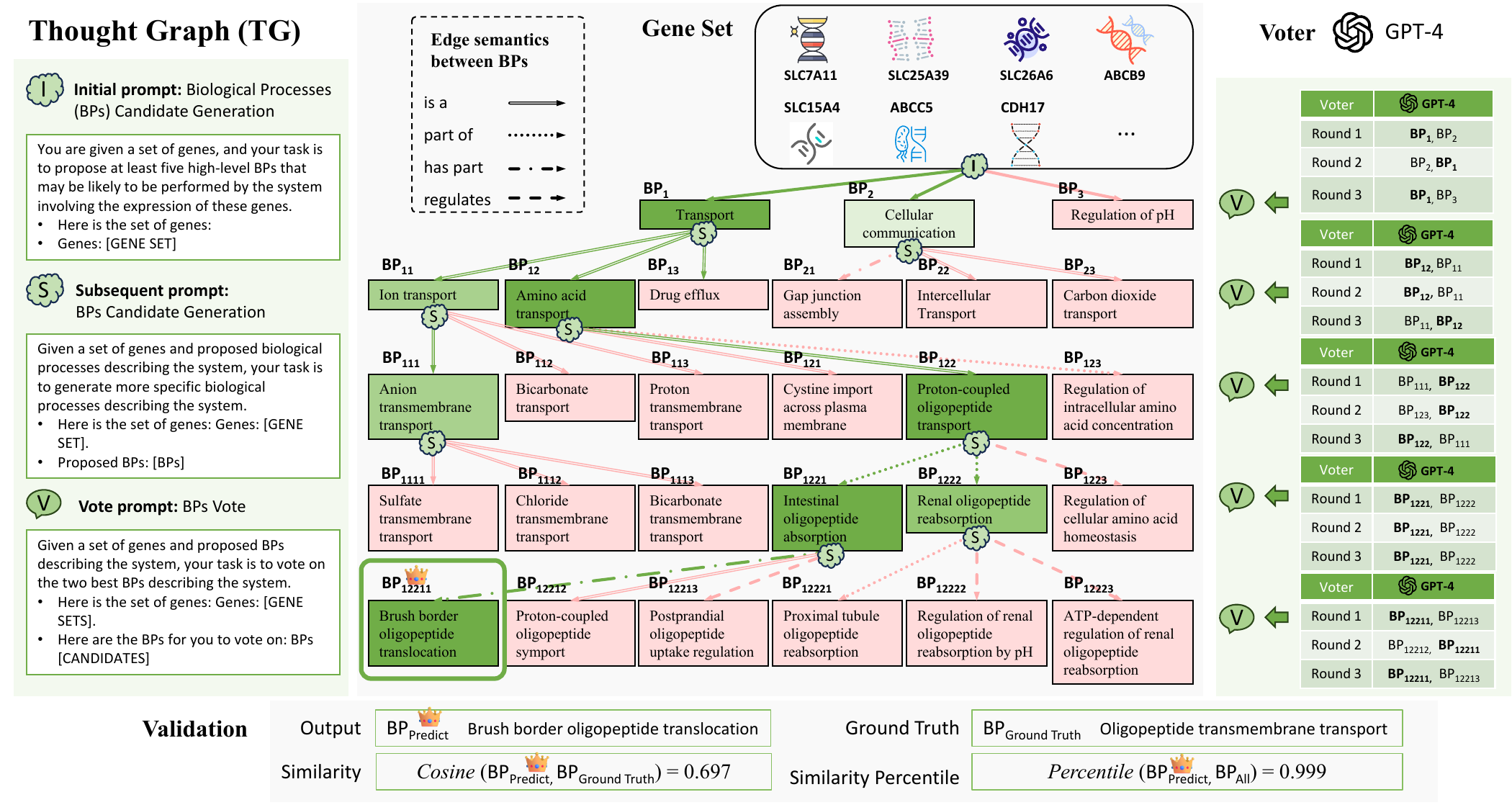}
    \caption{The flowchart presents the application of the \system to the Gene Ontology (GO) database. First, \system uses a gene set and initial prompt to generate three Biological Processes (BPs). Then, a voter evaluates and selects the best BP (dark green) and second best BP (light green), which are more accurately descriptive of the gene set. Each chosen BP, along with a subsequent prompt, is utilized to generate two additional, more specific BPs. This procedure is conducted recursively until \bptree has reached five layers. Finally, a voter chooses the final answer from the last layer.} 
    \label{fig:example}
\end{figure*}
\vspace{-0.3cm}
\section{Introduction}
The systematic study of human disease necessitates an in-depth understanding of the links between diseases, drugs, phenotypes, genes, and biological processes~\cite{chandak2023building}. Analyzing gene sets that share common biological functions, locations, or regulatory mechanisms can reveal patterns in gene behavior across health and disease states, contributing to the advancement of precision medicine for cancer treatment~\cite{doi:10.1073/pnas.0506580102}. Yet, the task of identifying biological processes from gene sets is fraught with challenges. Individual genes often display weak signals, and when strong signals are present, they rarely converge on a singular biological theme~\cite{doi:10.1073/pnas.0506580102}. This complexity is compounded when different research groups studying the same biological systems arrive at vastly divergent conclusions.

In response to these challenges, our paper introduces the \textbf{Thought Graph} framework that aims to address two critical aspects: firstly, 
it adopted a Tree-of-Thought (ToT) \cite{yao2023tree} architecture to facilitate thought expansion with Large Language Model (LLM), ensuring inclusive yet precise coverage of biological processes across varying specificity levels. 
Thought expansion is strategically directed with the assistance of a voter LLM, which guides the decision-making for future steps. This design aims to mitigate the potential discrepancies in human annotations encountered by researchers, yet ensure the quality of the generated processes.
Second, our framework prioritizes the integration of domain-specific external knowledge bases to understand the semantics of connections within the Thought Graph. Consequently, it creates semantic relationships like ``is-a'' and ``part-of'' among various thought steps. This strategy not only facilitates complex decision-making processes but also ensures a more nuanced and interconnected understanding of biological systems, facilitating data interoperability and knowledge integration.
Our novel contributions can be summarized as follows:
\begin{enumerate}
    \item We propose Thought Graph as a complex reasoning framework that generates diverse yet precise entities to tackle potential annotations discrepancies in biological processes.
    \item Thought Graph can generate thought graphs with edge semantics by recalling external knowledge (e.g., Gene Ontology) to build rich semantics among thought steps.
    \item We have successfully applied Thought Graph in biological process generation with significant improvement compared to SOTA methods, surpassing GSEA by 40.28\% and LLM baselines by 5.38\% in cosine similarity score, and identified the optimal steps of complex reasoning by balancing specificity and accuracy.
    % \item We have successfully applied \system in biological process generation with noteworthy improvement compared to SOTA methods, and identified the optimal steps of complex reasoning by balancing specificity and accuracy.
\end{enumerate}

\section{Related Work}
\subsection{LLM Reasoning} Prompt strategies attempt to decompose a complicated problem into a sequence of smaller sub-problems so that the problem becomes more manageable~\cite{zhou2023leasttomost}. One popular line of study is the Chain-of-Thought (CoT) \cite{wei2023chainofthought} series, structuring prompts to encourage the LLM to step through its reasoning process, such as Least-to-Most prompting~\cite{zhou2023leasttomost}, and Self-Consistency with CoT (CoT-SC)~\cite{wang2023selfconsistency}. However, these prompting strategies only utilize linear reasoning paths and struggle in tasks that require exploration and strategic lookahead. Alternatively, Tree of Thoughts (ToT) ~\cite{yao2023tree} and Graph of Thoughts (GoT) ~\cite{besta2023graph} excel in these sorts of tasks. LLM-based prompting frameworks' effectiveness is hindered by inherent limitations such as self-bias and hallucination. To address this, through in-context learning, our work introduces the semantics of edges within our Thought Graph, offering structural information. 
% for decision-making. 
% that assists LLMs in making more informed decisions.
\vspace{-0.2cm}
\subsection{Knowledge Graph for LLM Reasoning} 
LLMs exhibit limitations in integrating new knowledge and occasionally generate hallucinations. A survey ~\cite{agrawal2023knowledge} on knowledge-graph-based knowledge augmentation in LLMs reveals using knowledge graphs (KGs) as a source of external information has promising results in reducing hallucinations. For example, MindMap~\cite{wen2023mindmap} has developed a prompt pipeline enabling LLMs to comprehend and integrate KG input with their implicit knowledge. In our approach, we give LLM 
% some biological process 
examples from the gene ontology knowledge graph to enable 
% LLM to learn 
the edge semantics. 
% for the TG, 
% thereby providing a domain-specific context to improve the reasoning performance.
\subsection{LLM Reasoning in Biomedical Domain}
With the rise of LLMs, recent studies explore LLMs' application in various biomedical tasks. The gene set biological process was formulated by ~\cite{hu2023evaluation} as inputting a gene set to an LLM and outputting a biological process name that is predominant in the system and correctly describing the function of the gene set. It's challenging because it requires the LLM to accurately understand and interpret complex biological concepts, including the nuanced roles of genes in various cellular contexts and their interactions within intricate biological networks. Although their results ~\cite{hu2023evaluation} have shown that GPT-4 provides better biological process names than the conventional Gene Set Enrichment Analysis (GSEA)~\cite{doi:10.1073/pnas.0506580102}, the performance is still far from perfect. 
\section{Methodology}
% \begin{figure}[t]
%     \includegraphics[width=\linewidth]{images/workflow.pdf}
%     \centering
% 	 \caption{} 
% % 	 \vspace{-1.0pc}
%     \label{fig:workflow}
% \end{figure}

% \ethan{five steps}
% \ethan{We use $p\theta$ to denote a pre-trained LM
% with parameters $\theta$}

\subsection{Problem Formulation}
Given a gene set $X = \{x_1, x_2, ..., x_n\}$, where each $x_i$ is a gene, the objective $G = F(X)$ is to design a framework $F$ to generate a tree structure graph $G = (N, E)$ that represents the terms (e.g., biological processes or pathways) associated with the genes in $X$. In this graph, $N$ is the set of nodes, and $E$ is the set of edges between these nodes.

\subsection{Infrastructure of \system}
Our framework \system adapts ToT \cite{yao2023tree} as a graph generator to generate a curated tree graph $G$, named \bptree. \bptree contains terms as the nodes $N$ and their dependencies as edges $E$. ToT uses self-reflection to prune and only explore relevant paths. The result, after exploration, is a graph \bptree that illustrates the reasoning path and a final answer chosen selected from the last layer of the graph as the term that best describes the gene set $X = \{x_1, x_2, ..., x_m\}$. 

% In Tree-of-thought, at each step, a question $Q$ and a set of options $op = {o_1, o_2,..., o_k}$ are generated by an LLM. Multiple evaluation LLM agents $A = {A_1,...,A_k}$ would discuss each options and choose two correct options from $op$.  

\subsubsection{Thoughts expansion}
% \subsubsubsection{Method}
% The
% Given the full biological processes ontology graph $B,$ our goal is to generate a subgraph of the ontology $G \subset B$ that describes the set of genes  and $B$ share the unique root node "Biological Process," representing that both graphs lie in the biological processes branch of the complete gene ontology.

\system process with $n$ steps proceeds in a breadth-first fashion to generate a tree of depth $n$. At each step, the process expands the tree by generating a set of candidate nodes. The first step generates a set of general ``high-level'' terms that describe the gene set, and subsequent steps iterate on the candidate terms by proposing more specific but related terms. % that describe the gene set.

\textbf{Step 1 (Initial Expansion).} The first step is unique from all subsequent steps because its task is to generate the initial set of $k$ candidate terms $T_{i} = {t^1_1, \dots, t^1_k}$, where $t^i_j$ denotes the term $j$ from layer $i$. This set of candidate terms is generated with an ``initial prompt'' that takes the gene set as input: $T_{i} \sim p_\theta(t^{1}_{1 \dots k} \mid x_1 \dots, x_n)$. 

% \texttt{You are given a set of genes, and your task is to propose three high-level terms that may be likely to be performed by the system involving the expression of these genes. Genes: [gene1, gene2...]}. 

\textbf{Subsequent Steps (Recursive Expansion).} In step $i$, we use a Voter ($V$) to examine and vote across the candidate terms $T_{i}$: $V(p_\theta, T_{i})(t^{i}_{j}) = 1[t^{i}_{j}=t^{i}_{j}*]$, where a good term $t^{i}_{j}* \sim p^{vote}_\theta(t^{i}_{j}*| B)$ is based on comparing the candidate terms $T_{i}$ in the vote prompt, and select two best terms. 
%An example of the vote prompt is shown in Fig. 1. 
For each selected term from the previous step, $t^{i-1}_j$ and gene set $X$ are added to the ``subsequent prompt'' for the LLM generates $k$ new terms: 
$\{t^{i}_{1}, \dots, t^{i}_{k}\} \sim p_\theta(t^{i}_{1 \dots k} \mid x_{1 \dots n}, t^{i-1}_j)$. This process will be conducted recursively for $n-1$ times (minus the initial expansion). For the final layer,  $t^{n}_1 \sim t^{n}_k$ are presented to the LLM to choose the final answer.

% \texttt{Given a set of genes and proposed terms describing the system, your task is to generate more specific terms describing the system. Selected terms: [term1, term2]. Genes: [gene1, gene2...]}. 
% \ethan{prompt example}

% \textbf{Voter.} We modified the voter from ToT \cite{yao2023tree}'s state evaluator to examine the candidate terms. 
% We leverage two types of voters, a GPT-4 model and a GPT model augmented with a GProfiler, a popular gene set analysis tool. 

% An example of the vote prompt is shown in Appendix A. 

% The second voter designs to mitigate the limitation of GPT-4 of not accessing external tools or web pages. Thus, we also incorporate GProfiler with GPT-4. GProfiler identifies a table of biological terms significantly associated with a given list of genes. We filter the results by finding the five most similar biological terms to each candidate, based on their cosine similarity using SapBERT embeddings \cite{liu2021selfalignment}. Then, the filtered results ([TABLE]) are added to which the vote prompt includes GProfiler's analysis; an example is shown as the following: 

\subsection{\bptree}
The \bptree output provides a representation of the step-wise reasoning process and integrates edge and node semantics for domain-specific context. Each node $n_i \in N$ is a unique biological process, arranged hierarchically to reflect varying levels of specificity. The edges $E$ represent the relationships between these processes. Specifically, we use four pre-defined relations from the Gene Ontology (GO): \textit{is a}, \textit{part of}, \textit{has part}, and \textit{regulates.} These relations establish a hierarchy where, for instance, if A \textit{is a} subtype of B, A is deemed more specific than B. This approach helps to elucidate the nuanced relationships between different biological processes, as detailed in the GO database.\footnote{https://geneontology.org/docs/ontology-relations/}

\section{Experiment \& Evaluation}

\subsection{Data Collection}
The GO database \cite{article} forms the basis of our study. We specifically use a dataset compiled by Hu et al. \cite{hu2023evaluation} from the Biological Process branch of Gene Ontology consisting of 12,214 human gene sets, each annotated with a biological process name and description. Due to constraints in financial and computational resources, we randomly select 100 samples from this dataset for evaluation.

\subsection{Baselines and Model Description}
Our evaluation framework includes one domain-specific tool and five LLM baselines. GSEA (gene set enrichment analysis) \cite{doi:10.1073/pnas.0506580102} is a statistical method for associating the expression of groups of genes with biological processes. Our LLM baselines involve different approaches. Input-Output (IO) Prompting with zero-shot and zero-shot-9 prompts generate one and nine unique terms for a single gene set, respectively, with no examples, while few-shot includes five question-answer examples. Chain-of-Thought (CoT) employs the two top pathways from \system for detailed step-by-step prompting. The approach by Hu et al. \cite{hu2023evaluation} integrates expert-curated prompts with specific guidelines that solicit post-hoc critical analysis. For all LLM instances, we use GPT-4 (\textit{gpt-4-1106-preview}) in Chat Completion mode with temperature 0.7. In Thought Graph, we set the number of steps to five and vote on two samples at each step to proceed.

%, to guide the decision-making process.

\subsection{Evaluation Methods}
We use two evaluation metrics: cosine similarity and similarity percentile. Cosine similarity measures the semantic similarity of the predicted term to ground-truth term from 0 (no similarity) to 1 (identical). We calculate similarity using embeddings from SapBERT \cite{liu2021selfalignment}, a masked language model trained to model medical entity relations. After calculating the similarity between the predicted and ground-truth terms, we also calculate the similarity between the predicted term and all 12,214 terms in our dataset to form a null distribution. The percentile score is the percentile of the similarity between the predicted term and the ground-truth term in our null distribution. We also include the proportion of similarity percentiles greater than 99\% as a proxy for accuracy.

Among the nine nodes that receive positive votes (indicated as green nodes in Fig. \ref{fig:example}), the one with the highest similarity score is selected as the best score (b), while the score of the node predicted by \system is recorded as the predicted score (p). To establish a fair baseline comparison, we implemented IO zero-shot-9 to generate nine answers, and select the best of these for evaluation.
% In evaluating \system, the cosine similarity of each node in \bptree to the ground truth is computed. Among voted nodes (green nodes in Fig. \ref{fig:example}), the node with the highest similarity score is designated as the best score (b), and the score of the node predicted by \system is noted as the predicted score (p). 
% The zero-shot-9 (b) is derived from the answer with the highest similarity score.

\begin{figure}[t]
    \includegraphics[width=\linewidth]{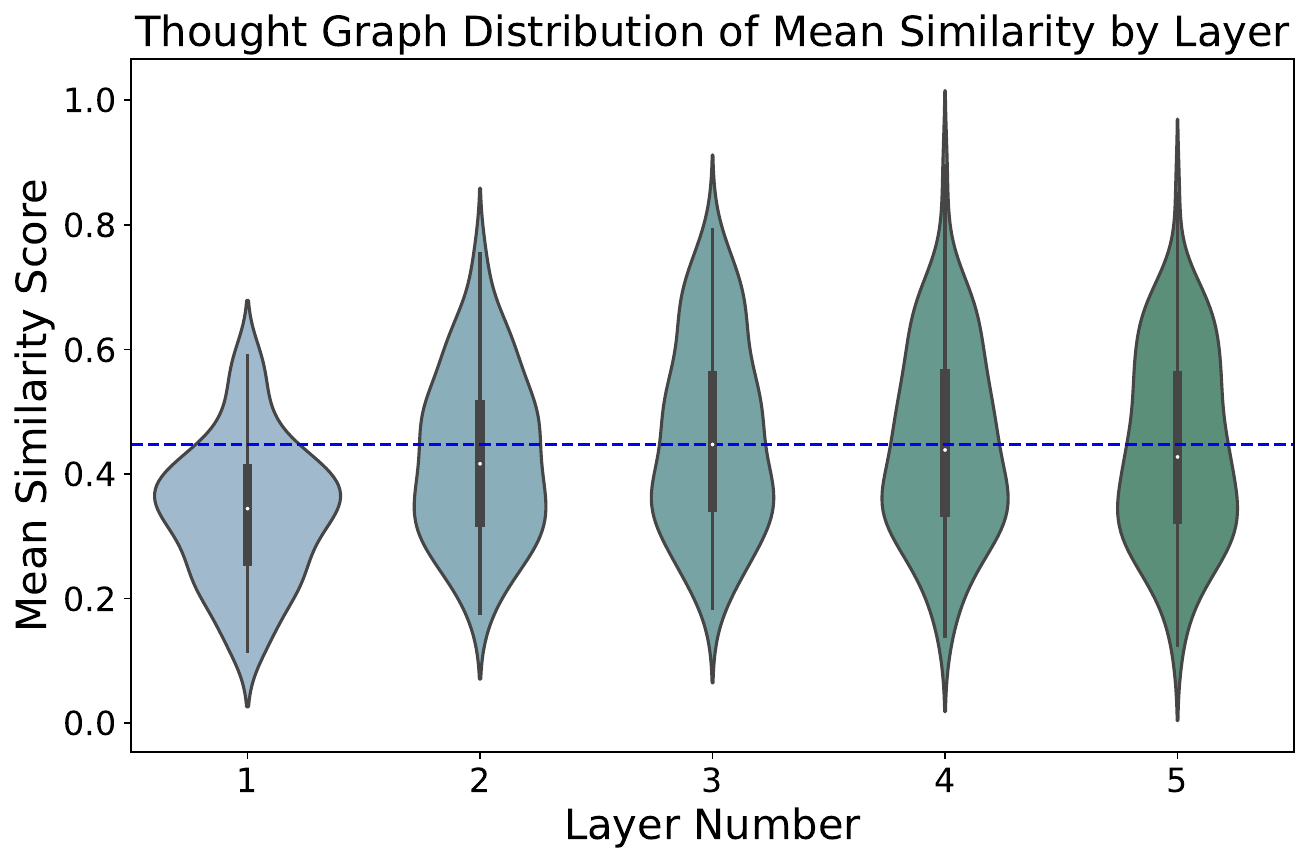}
    \centering
	 \caption{The distribution of the mean similarity score at each layer using \system (p). The blue line denotes the median of layer 3.} 

    \label{fig:layer_boxplot}
\end{figure}
% \vspace{-0.5cm}

% \vspace{-0.2cm}

\begin{table}[h]
\centering
\begin{tabular}{|l|c|c|c|}
% \hline
% Domain-specific tool  & Cosine Similarity & Similarity Quantile  \\
% \hline

\hline
Method & Similarity & Percentile & Percentile > 99\% \\
\hline
GSEA & 24.78\% & 52.00\% & 17\% \\
\hline
IO zero-shot & 45.75\% & 77.00\% & 27\% \\
\hline
IO zero-shot-9 (b) & 59.68\% & 91.42\% & 61\% \\
\hline
IO few-shot & 48.73\% & 81.85\% & 32\% \\
\hline
CoT & 28.83\% & 43.71\% & 0\% \\
\hline
Hu et al. \cite{hu2023evaluation} & 52.31\% & 84.44\% & 43\% \\
\hline
\system (p) & 48.53\% & 80.90\% & 42\%\\
\hline
\system (b) & \textbf{65.06\%} & \textbf{95.05}\% & \textbf{65\%} \\
\hline
\end{tabular}
\caption{Mean cosine similarity, mean cosine similarity percentile, and proportion percentile above 99\% of a domain-specific tool and seven LLM methods on 100 GO data samples.}
\vspace{-0.8cm}
\label{table:performance}
\end{table}

% \hline
% Method  & Cosine Similarity & Similarity Percentile  \\
% \hline
% GSEA & 22.10\% & 45.70\%  \\
% \hline
% zero-shot & 45.75\% & 77.00\% \\
% \hline
% zero-shot-9 (b) & 59.68\% & 91.42\% \\
% \hline
% few-shots & 48.73\% & 81.85\% \\
% \hline
% CoT & 28.83\% & 43.71\% \\
% \hline
% Hu et al. \cite{hu2023evaluation} & 52.31\% & 84.44\% \\
% \hline
% \system (p) & 48.53\% & 80.90\% \\
% \hline
% \system (b) & \textbf{65.06\%} & \textbf{95.05}\% \\
% \hline
% \system (b) & \textbf{71.48\%} & \textbf{97.50}\% \\
% \hline
% \system+ GP (p) & 44.05\% & 76.65\% \\
% \hline
% \system+ GP (b) & 67.18\% & \textbf{97.15\%} \\
% % \hline
% \end{tabular}
% \caption{Performance of domain-specific tool and models on 100 GO data samples.}
% \label{table:performance}
% \end{table}

% ``Cosine Similarity'' denotes the cosine similarity between generated text and ground truth; ``similarity percentile'' score denotes the similarity score against all biological processes in the GO database.

\subsection{Performance Evaluation}
\textbf{Overall Performance:} Table \ref{table:performance} indicates that \system (b) achieves the top performance in both cosine similarity (65.06\%) and similarity percentile (95.05\%). In particular, we want to posit that IO zero-shot learning emphasizes coverage across a wide range of biological process names (diversity), while the CoT focuses on an in-depth exploration of these names (specificity), whereas our framework is designed to balance both. \system (b) outperforms IO zero-shot-9 (b) and CoT, indicating that depth without breadth, or vice versa, is insufficient. 
% \system and other LLM baselines outperform GSEA, and we also noticed GSEA cannot provide any terms for 26\% of the time, highlighting the advantage of the LLMs. 
% However, that \system (b) outperforms all baselines, including zero-shot-9, assures us that our approach to generating candidate sets of terms is promising, and that it is adept at generating a correct answer, but further optimization is needed to select it from the graph. 
% Table \ref{table:performance} indicates that \system (b) achieves the top performance in both cosine similarity (65.06\%) and similarity percentile (95.05\%). 
% In particular, we note that \system (b) outperforms zero-shot-9 (b), suggesting that \system benefits from our specificity framework design. Interestingly, CoT's performance is lower than both zero-shot and few-shot, indicating that depth without breadth can negatively impact results. 
\system and other LLM baselines outperform GSEA, and we also noticed GSEA cannot provide any terms for 26\% of the time, highlighting the advantage of the LLMs. In addition, \system (p) scores lower than few-shot and Hu et al. baselines. This may be the result of our decision to constrain the final answer to the last layer. However, that \system (b) outperforms all baselines, including zero-shot-9, assures us that our approach to generating candidate sets of terms is promising, and that it is adept at generating a correct answer, but further optimization is needed.

%The \system framework, however, is optimized to generate a candidate set of terms at varying levels of specificity with legible relations between them, making no sacrifices along the bias-variance trade-off. 

% another baseline generating diverse answers, \system significantly excels by 5.38\%, underscoring the value of deliberate reasoning and specificity. Interestingly, CoT's performance is lower than both zero-shot and few-shot, indicating that depth without breadth can negatively impact results. We also noticed GSEA cannot provide any terms for 26\% of the time, highlighting the advantage of the LLMs. 
% % The study by Hu et al. \cite{hu2023evaluation} over zero-shot further emphasizes the importance of expert-curated prompts. 
% Nonetheless, \system (p) shows that our framework faces challenges in identifying the final terms; this can be due to our framework constraining output to the fifth layer; further investigation is needed in dynamic path-finding studies.
% %\ethan{our nine is more powerful than the other}

\textbf{\bptree Analysis:} Layer-by-layer analysis in Fig.\ref{fig:layer_boxplot} demonstrates increasing performance from layers 1 to 3, followed by a decrease in layers 4 and 5. This trend suggests a trade-off between specificity and accuracy, with layer 3 the optimal level by a small margin. While the performance at layer 1 is lower, this is largely because our initial prompt specifically requests ``high-level'' terms, and only generates three of them. As expected, the variance in mean similarity scores increases with the number of layers, as deeper layers explore deeper and more distant parts of the ontology, but stabilize after layer 3. In the latter layers, more specific terms are often voted out in favor of more accurate, general terms, demonstrating the ability of the voting mechanism to dynamically moderate specificity. Though our results reflect a modest sample size, layer 3 emerges as an early candidate for the optimal depth.

\vspace{-0.2cm}
\section{Conclusion}
\system represents an advancement in the field of gene ontology and bioinformatics. Integrating gene set analysis with semantic graphs allows for a more nuanced and comprehensive understanding of biological processes. The effectiveness of the \bptree in mapping complex gene interactions and functions has been demonstrated, showing its potential to outperform existing methods. This novel method not only enhances the accuracy of gene set analysis but also opens avenues for research in understanding genetic influences on various BPs. Future work can expand on this foundation, exploring broader applications and measuring uncertainty in complex reasoning.
% Future work can expand on this foundation and explore broader applications. % and measuring uncertainty in complex reasoning. %for even more detailed insights.

% \vspace{-0.3cm}
\vspace{-0.2cm}
\section{Acknowledgement}
We thank the support from NIH (OTA-21-008, R01LM014306-01) and NSF (NSF 2303038, NSF 2333703).

\vspace{-0.15cm}

%%
%% The acknowledgments section is defined using the "acks" environment
%% (and NOT an unnumbered section). This ensures the proper
%% identification of the section in the article metadata, and the
%% consistent spelling of the heading.

% \begin{acks}
% To Robert, for the bagels and explaining CMYK and color spaces.
% \end{acks}

%%
%% The next two lines define the bibliography style to be used, and
%% the bibliography file.
% \vspace{-7pt}
\bibliographystyle{ACM-Reference-Format}
\bibliography{bib}

%%
%% If your work has an appendix, this is the place to put it.
\appendix

\end{document}